 \documentclass[tablecaption=bottom,wcp]{jmlr} 

 \usepackage{dsfont}

\usepackage{booktabs}
\usepackage[load-configurations=version-1]{siunitx} 

\theorembodyfont{\upshape}
\theoremheaderfont{\scshape}
\theorempostheader{:}
\theoremsep{\newline}

\jmlrproceedings{}{}

\newcommand{\x}{\mathbf{x}}
\newcommand{\z}{\mathbf{z}}
\newcommand*\compactparagraph[1]{\textbf{#1} \quad}
\newcommand{\appropto}{\mathrel{\vcenter{
  \offinterlineskip\halign{\hfil$##$\cr
    \propto\cr\noalign{\kern2pt}\sim\cr\noalign{\kern-2pt}}}}}

\title[Bootstrap Your Flow]{Bootstrap Your Flow}

\author{\Name{Anonymous Authors}\\
  \addr Anonymous Institution}

  \author{\Name{Laurence Illing Midgley} \Email{laurencemidgley@gmail.com}\\
   \Name{Vincent Stimper} \Email{vs488@cam.ac.uk}\\
   \Name{Gregor N.\ C.\ Simm} \Email{gncs2@cam.ac.uk}\\
   \Name{José Miguel Hernández-Lobato} \Email{jmh233@cam.ac.uk}\\
\addr Department of Engineering, University of Cambridge}

\begin{document}

\maketitle

\begin{abstract}
Normalizing flows are flexible, parameterized distributions that can be used to approximate expectations from intractable distributions via importance sampling. 
However, current flow-based approaches are limited on challenging targets where they either suffer from mode seeking behaviour or high variance in the training loss, or rely on samples from the target distribution, which may not be available.
To address these challenges, we combine flows with annealed importance sampling (AIS), while using the $\alpha$-divergence as our objective, in a novel training procedure, FAB (Flow AIS Bootstrap). 
Thereby, the flow and AIS improve each other in a bootstrapping manner.
We demonstrate that FAB can be used to produce accurate approximations to complex target distributions, including Boltzmann distributions, in problems where previous flow-based methods fail. 
We provide code for our experiments at \url{https://github.com/lollcat/FAB-2021}.


\end{abstract}

\section{Introduction}

Estimating expectations with respect to target distributions that cannot be sampled from is a challenging task with many real world applications, 
such as estimating equilibrium properties of physical systems governed by the Boltzmann distribution \citep{Lelievre2010}.
Boltzmann generators \citep{noe2019boltzmann}, which use normalizing flows to approximate the Boltzmann distribution, are a recent approach with growing interest \citep{dibak2020,kohler2021}. 
For challenging problems, current approaches to training Boltzmann generators rely partly on samples from the target for training by the maximum likelihood  \cite{wu2020stochasticNF}.
Such samples are obtained through expensive Molecular Dynamics simulations \citep{leimkuhler2015}.
Although flows can be trained without samples from the target, current methods for this suffer from either being mode seeking or high variance in the loss, which leads to inferior performance on challenging problems \citep{stimper2021}.

To address these challenges, we propose using the $\alpha$-divergence with $\alpha=2$ as the training objective, which is  mass covering,
and employ annealed importance sampling (AIS) to bring the samples from the flow model closer to the target, reducing variance in the objective.
In our experiments, we apply our method, \textbf{F}low \textbf{A}IS \textbf{B}ootstrap (\textbf{FAB}), to a 2D Gaussian mixture distribution 
as well as to the ``Many Well'' problem and show that it outperforms competing learning algorithms.

Although we focus on a toy problem and the Boltzmann distribution, our approach for training flows to approximate intractable distributions is not specific to these problems, and can be applied to any target distribution for which an un-normalized probability density function is defined.
One key application of interest that we leave for future work is approximate inference over Bayesian posterior distributions (for example the posterior of a Bayesian Neural Network). 
In the context of Bayesian inference, in terms of computational complexity, FAB lies between Markov chain Monte Carlo, the computationally expensive gold-standard for Bayesian inference, and variational methods such as Bayes-by-backprop \citep{blundell2015weight} which provide a computationally cheaper alternative, but suffer from worse performance. 

\section{Background}

\compactparagraph{Normalizing flows}
Given a random variable $\z$ with distribution $q(\z)$ a normalizing flow \citep{rezende2015variational,earlyFlowPaperTabak} uses an invertible map \mbox{$F: \mathds{R}^d \rightarrow \mathds{R}^d$} to transform $\z$ yielding the random variable $\x = F(\z)$ with the distribution
\begin{equation}
    q(\x) = q(\z) \left| \det(J_{F}(\z))\right| ^{-1}.
\end{equation}
If we parameterize $F$ we can use this as a model to approximate a target distribution $p$. 
If the target density $p(\x)$ is available, the flow is usually trained by minimizing the KL divergence, which is estimated via Monte Carlo with samples from the flow model. Alternatively, we could use the $\alpha$-divergence \citep{zhu1995information}, which is defined by 
\begin{equation}
\label{eqn:alpha-divergence}
    D_{\alpha}(p \| q)=\frac{\int_{x} \alpha p(\mathbf{x})+(1-\alpha) q(\mathbf{x})-p(\mathbf{x})^{\alpha} q(\mathbf{x})^{1-\alpha} d \x}{\alpha(1-\alpha)},
\end{equation}
as an objective \citep{campbell2021gradient,muller2019}. If $\alpha=2$, minimizing it corresponds to minimizing the variance of the importance weights $w_\text{IS}(\x) = \frac{p(\x)}{q(\x)}$.
In contrast to KL divergence, which is mode seeking, $D_{\alpha=2}(p \| q)$ is mass covering which is more desirable when approximating multimodal targets.
In this case, the $\alpha$-divergence can be rewritten as
\begin{equation}
    D_{\alpha=2}(p \| q) \propto  \int \frac{p(\mathbf{x})^{2}}{q(\mathbf{x})} d \mathbf{x}
    = \operatorname{E}_{q(\mathbf{x})} \left[  w_\text{IS}(\x) ^ 2  \right]
    = \operatorname{E}_{p(\mathbf{x})} \left[  w_\text{IS}(\x)  \right].
    \label{eqn:alpha_2_over_p_vs_q}
\end{equation}
This objective can be estimated either with samples from $p(\x)$ or $q(\x)$.
Since the integral is dominated by regions with high $p(\x)$ and low $q(\x)$, the estimate will exhibit higher variance if we sample from $q$ than if we sample from $p$.

\compactparagraph{Annealed importance sampling}
AIS begins by sampling from an initial proposal distribution $\x_0 \sim p_0 = q$, being the flow in our case, and then transitions via MCMC through a sequence of intermediate distributions, $p_1$ to $p_{N-1}$, to produce samples $\mathbf{x}_{N-1}$ closer to the target distribution $p_N = p$ \citep{neal2001annealed}.
Each transition $T_j$ is a Markov chain that leaves the intermediate distribution $p_j$ invariant.
AIS conventionally returns the importance weights for the samples, which are in the form
\begin{equation}
w_\text{AIS}(\mathbf{x}^{(i)})=\frac{\tilde{p}_{1}\left(\mathbf{x}_{0}\right)}{p_{0}\left(\mathbf{x}_{0}\right)} \frac{\tilde{p}_{2}\left(\mathbf{x}_{1}\right)}{\tilde{p}_{1}\left(\mathbf{x}_{1}\right)} \cdots \frac{\tilde{p}_{N-1}\left(\mathbf{x}_{N-2}\right)}{\tilde{p}_{N-2}\left(\mathbf{x}_{N-2}\right)} \frac{\tilde{p}_{N}\left(\mathbf{x}_{N-1}\right)}{\tilde{p}_{N-1}\left(\mathbf{x}_{N-1}\right)}
\label{equ:wais}
\end{equation}
where $\mathbf{x}^{(i)} = \mathbf{x}_{N-1}$
and we indicate that the probability density functions may be unnormalized with $\tilde{p}$. They exhibit variance reduction compared to their counterparts $w_\text{IS}(\x)$.
Hamiltonian Monte Carlo (HMC) provides a suitable choice of transition operator for challenging problems \citep{neal1995bayesian}.

\section{Normalizing Flow Annealed Importance Sampling Bootstrap}
FAB, defined in Algorithm \ref{algorithm:FAB}, uses the $\log$ of a Monte Carlo estimate of $(D_{\alpha=2}(p \| q))$ as a training objective. 
Furthermore, we introduce AIS into the training loop, improving the gradient estimator for minimising $D_{\alpha=2}(p \| q)$ by writing the loss function to train the flow as an expectation over $p(\mathbf{x})$, and estimating it with the samples and importance weights generated by AIS with the flow as the initial distribution. 
If we plug in \eqref{equ:wais} in \eqref{eqn:alpha_2_over_p_vs_q}, compute the expectation over $p(\x)$ through AIS and take the $\log$ of this, we obtain our loss
\begin{equation}
    \mathcal{L} (\theta)=
    \log \sum_{l=1}^L \exp \left( \log  \bar{w}^{(l)}_{\text{AIS}} + \left(\log p(\bar{\x}^{(l)}_{\text{AIS}}) - \log q(\bar{\x}^{(l)}_{\text{AIS}})\right) \right),
\end{equation}
where $\theta$ are the flow's parameter and $\bar{w}_{AIS}$ and $\bar{\x}_{AIS}$ have blocked gradients\footnote{We only want to calculate the gradient with respect to the $\log(D_{\alpha=2}(p \| q))$, and not with respect to the importance weights and samples generated by AIS, which are also a function of the flow's parameters.}. 
The introduction of the $\log$ reduces the variance in the loss but adds bias (as a result of Jensen's inequality) which decreases as $L$ increases.
This is similar to the loss used for black box $\alpha$-divergence minimisation in \cite{hernandez2016black}.

This method obtains the benefit of bootstrapping, where AIS is used to improve the flow by improved estimation of the loss's gradient, which is used to update the flow. This in turn improves AIS by improving its initial distribution. 
The effective sample size of the trained flow model, which may be limited e.g. through expressiveness constraints of the specific flow, can also be improved by using AIS after training. 

\begin{algorithm2e}[htbp]
\DontPrintSemicolon
Set target $p$ \;
Initialise proposal $q$ parameterized by $\theta$ \;
\For{\text{iteration} = 1, $M$}{
    Sample batch $\mathbf{x}_q^{(1:L)},\log q(\mathbf{x}_q^{(1:L)})$ from $q$ \;
    Generate batch $\mathbf{x}_{AIS}^{(1:L)}$, $\log w_{AIS}^{(1:L)}$ from AIS seeded with  $\mathbf{x}_q^{(1:L)},\log q(\mathbf{x}_q^{(1:L)})$ \;
    Calculate FAB loss: $\mathcal{L} (\theta)$\;
    Perform gradient descent on $\mathcal{L} (\theta )$\;
    }
\caption{FAB for minimisation of $D_{\alpha=2}(p \| q) $}
\label{algorithm:FAB}
\end{algorithm2e}

\section{Experiments}
\subsection{Mixture of Gaussians Problem}
\label{sec:mog}
We begin with a simple two dimensional mixture of Gaussians target distribution. 
To estimate expectations $\mathbb{E}_{p(\mathbf{x})} \left[ f(\mathbf{x}) \right]$ with our proposal distribution, we set $f(\mathbf{x})$ to be the toy quadratic function
$
    f(\mathbf{x}) = \mathbf{a}^T \left(\mathbf{x} - 2 \mathbf{b} \right) + 2 \left(\mathbf{x} - 2\mathbf{b} \right)^T \mathbf{C} \left(\mathbf{x} - 2 \mathbf{b} \right),
$
where the elements of $\mathbf{a}$, $\mathbf{b}$ and $\mathbf{C}$ are sampled from a unit Gaussian and then fixed for the problem. 
This allows us to inspect the bias and variance of estimates of the expectation of this toy function, as a further measure of performance.
We compare FAB\footnote{For all experiments, we use FAB with only 2 intermediate AIS distributions. For the transition operator between AIS distributions we use HMC with 1 outer step and 5 inner steps. } to flows trained through minimisation of KLD\footnote{For brevity we refer to a flow/SNF trained through minimisation of KL divergence, estimated with samples from the flow/SNF, as simply being “trained with KLD”.},
as well as Stochastic Normalizing Flows (SNFs) \citep{wu2020stochasticNF}.
Like FAB, SNFs also combine flows with stochastic sampling methods such as MCMC, but instead focus on improving the flow's expressive power. The SNF is trained with the KL divergence as well. For all models we choose real NVP \citep{dinh2017RealNVP} as the flow architecture.

\begin{figure}[htbp]
    \centering
    \includegraphics[width=\textwidth]{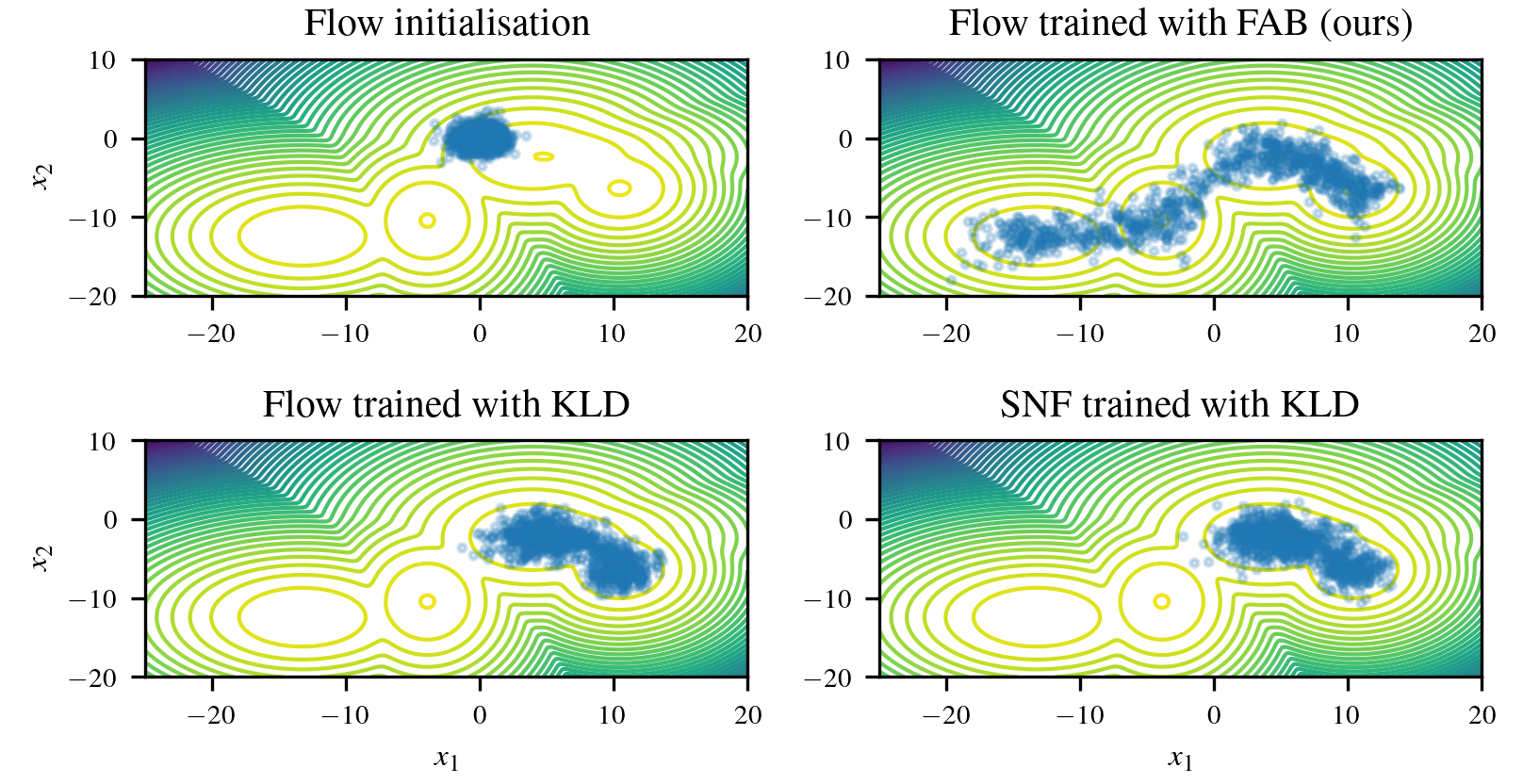}
    \caption{
     Mixture of Gaussians problem target probability contours, overlaid with samples from trained models.
    With FAB the flow captures all of the modes, while the flow and SNF trained with KLD fit a subset of the modes.}
    \label{fig:MoG}
\end{figure}

In Figure \ref{fig:MoG} we see that FAB allows us to train a flow that captures the shape of the target distribution well, while the flow and SNF trained with KLD both fail, capturing only a subset of the modes. 
Table \ref{table:MoG}\footnote{Effective sample size (ESS) for both experiments is calculated with $10^6$ samples.
Bias and standard deviation are calculated using 100 runs of 1000 samples.
We calculate the mean target log likelihood using 10000 samples.}
shows that with FAB the trained flow may be used for accurate computation of expectations with respect to the target,
while the alternative approaches yield highly biased estimates. 

\begin{table}[htbp]
  \caption{
  Performance of FAB vs a flow and SNF trained with KLD on the Mixture of Gaussians problem.
  For FAB all metrics are for the trained flow, while the metrics after AIS are provided in brackets where applicable.}
  \label{table:MoG}
  \centering
  \begin{tabular}{lcccc}
    \toprule
    Model     & Mean $\log q (\mathbf{x})$, $\mathbf{x} \sim p(\mathbf{x})$    & ESS (\%) & Bias (\%) &  Std (\%) \\
    \midrule
    FAB & -5.2  & 70.1 (77.5)  & 1.2 (0.5) & 5.8 (5.5)  \\
    Flow trained with KLD  & -14.4  &  0.05  & 99.6 & 19.8 \\ 
    SNF trained with KLD & N/A & 0.02 & 104.2 & 9.7\\ 
    \bottomrule
  \end{tabular}
\end{table}

\subsection{The Many Well Problem}
\begin{figure}[htbp]
    \centering
    \includegraphics[width=\linewidth]{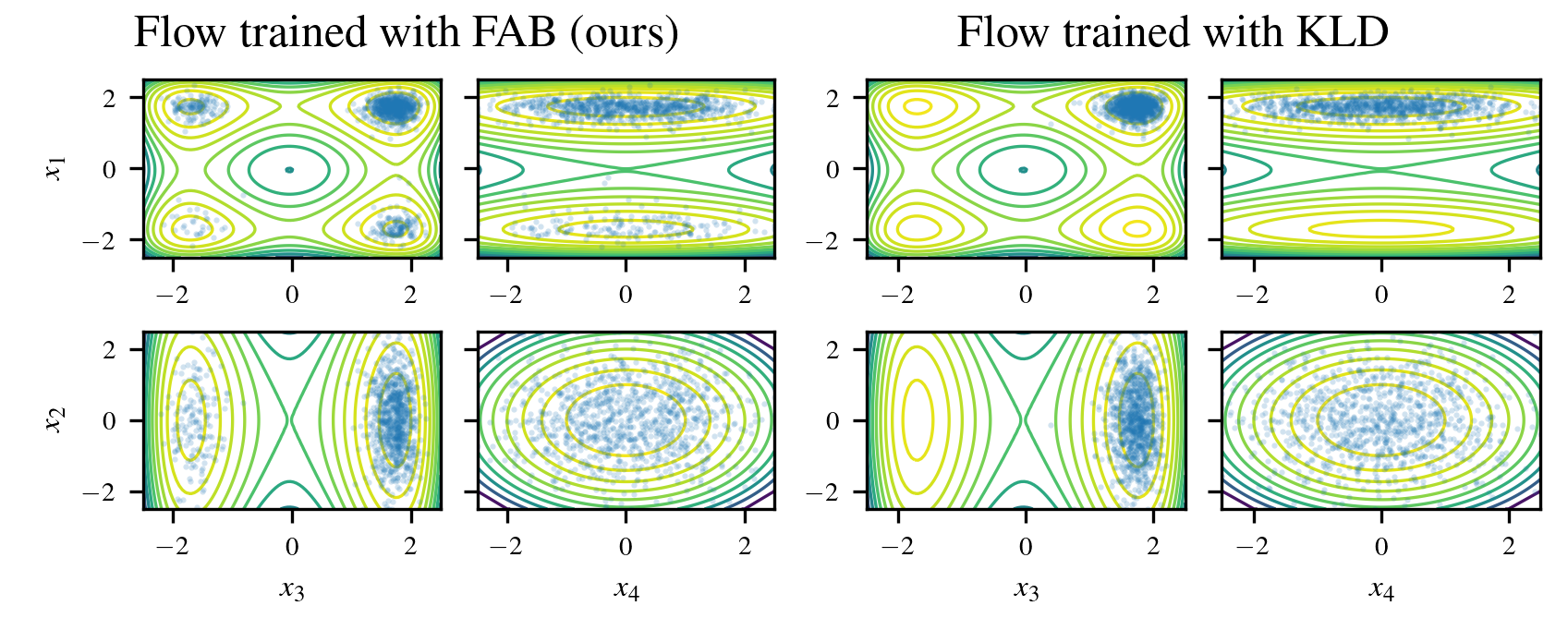}
    \caption{Target probability contours for pairs of marginal distributions for the first four dimensions of the 16 dimensional Many Well Problem, overlaid with samples from trained models.
    With FAB the flow captures all of the marginals' modes, while the flow trained with KLD, fits a subset of them.}
\label{fig:ManyWell}
\end{figure}

For a more challenging problem, we test FAB against a flow trained by KLD on a 16 dimensional “Many Well” problem, which we create by repeating the Double Well Boltzmann distribution from \citep{noe2019boltzmann} 8 times. 
We create a hand-crafted test-set for this problem where we place a point on each of the 256 modes of the target.
In Table \ref{table:many_well} we see that FAB allows us to train a model that has a far superior test set log-likelihood and ESS than training a flow with KLD. 
In Figure \ref{fig:ManyWell} where we visualise a subset of the marginal distributions for pairs of dimensions, we see that the flow trained with FAB has captured the shape of the target well, while the flow trained with KLD fits only a subset of the modes.

\begin{table}[htbp]
  \caption{Performance of FAB vs a flow trained by KLD on the Many Well Problem. Metrics are provided with respect to the trained flows, while for FAB, ESS after AIS is provided in brackets.}
  \label{table:many_well}
  \centering
  \begin{tabular}{lll}
    \toprule
    Model     & Test set mean $\log q (\mathbf{x})$    & ESS (\%) \\
    \midrule
    FAB & -14.5  & 79.6 (85.2)    \\
    Flow trained by KLD     & -86.2 & 0.01 \\
    \bottomrule
  \end{tabular}
\end{table}

\newpage
\section{Discussion}

In SNFs, MCMC is combined with normalizing flows by introducing sampling layers between the standard flow layers to improve the expressiveness of the flow \citep{nielsen2020,wu2020stochasticNF}. 
They are usually trained with samples from the target and perform poorly when trained exclusively with samples from the flow, see section \ref{sec:mog}. 
Instead, FAB focuses on improving the training procedure. 
However, the two contributions are orthogonal, so SNFs could be trained with FAB as well. Similarly, FAB works with other flow architectures \citep{chen2019,grcic2021} and base distributions \citep{papamakarios2017,stimper2021} than those we used.
In \cite{ding2019VAE-AIS} AIS is applied in the context of variational inference to improve estimation of the marginal likelihood gradient for varitational autoencoders' decoder. 
This approach may be extended by FAB to also improve the encoder training, through minimising the FAB loss with respect to the parameters of the encoder, with the latent posterior as the target distribution.
This may be advantageous relative to \citep{ding2019VAE-AIS}, as well as MCMC based approaches like \citep{hoffman2017MCMC-VAE} for training the decoder, because if the flow-based encoder can learn a good approximation of the latent posterior, then this alleviates the requirement to run MCMC for long in order to obtain good samples for the decoder. 
In neural transport MCMC \citep{hoffman2019neutra}, a flow is trained via KL divergence minimisation with a target distribution, and then is used to perform MCMC with the target in the latent space of the flow. 
This improves the geometry in which MCMC takes place, allowing for better mixing.
Neural transport MCMC relies on the flow capturing the geometry of the target distribution and can worsen sampling in the tails of the target distribution if the flow fails to capture the target well.
Thus by replacing the KL divergence training objective with the FAB training, our approach may be useful for improving the robustness of neural transport MCMC. 

We proposed FAB, a novel method of combing flows with AIS in a training procedure that allows them to improve each other in a bootstrapping manner.
For future work we hope to (1) identify how important each component of FAB is, such as the number of intermediate AIS distributions used and the form of the loss (2) perform a rigorous benchmarking of FAB's performance and computational complexity relative to alternative approaches such as MCMC, (3) explore the aforementioned connections of FAB to other recent advances in literature, and (4) scale up FAB and apply it to to more challenging real world problems, for example Boltzmann distributions of complex molecules. 


\begin{acks}
	GNCS and JMHL acknowledge support from a Turing AI Fellowship under grant EP/V023756/1.
	This work has been performed using resources operated by the University of Cambridge Research Computing Service,
	which is funded by the EPSRC (capital grant EP/P020259/1) and DiRAC funding from the STFC (\url{http://www.dirac.ac.uk/}). 
	LIM acknowledges support from the Cambridge Trust, the Skype Foundation, and the Oppenheimer Memorial Trust.
\end{acks}


\bibliography{references}

\appendix

\section{FAB derivation}\label{apd:first}
We aims to minimise
\begin{equation}
    D_{\alpha=2}(p \| q) \propto  \int \frac{p(\mathbf{x})^{2}}{q(\mathbf{x})} d \mathbf{x}
    = \operatorname{E}_{p(\mathbf{x})} \left[  \frac{p(\x)}{q(\x)}  \right].
    \label{eqn:alpha_2_over_p}
\end{equation}
We obtain a gradient estimator by differentiating Equation \ref{eqn:alpha_2_over_p} with respect to the parameters $\theta$ of the flow model
\begin{equation}
\label{eqn:gradient_estimator_alpha_div}
    \nabla_\theta \left[ \operatorname{E}_{p(\mathbf{x})} \left[  \frac{p(\mathbf{x})}{q_{\theta}(\mathbf{x})}\right] \right] =\operatorname{E}_{p(\mathbf{x})} \left[\nabla_\theta  \frac{p(\mathbf{x})}{q_{\theta}(\mathbf{x})}\right], 
\end{equation}
where we have used the fact that since $\theta$ is independent to samples from $p(\mathbf{x})$, we can move $\nabla_\theta$ inside the expectation. 
If we set 
$f(\mathbf{x}) = \nabla_\theta  \frac{p(\mathbf{x})}{q_{\theta}(\mathbf{x})}$,
we see that Equation \ref{eqn:gradient_estimator_alpha_div} is in the form $\mathbb{E}_{p(\mathbf{x})} \left[ f(\mathbf{x}) \right]$, so we can estimate it with AIS. 

With this goal in mind, during the training loop, we generate a batch of importance weights $w^{(1:L)}_{\text{AIS}}$, and samples $\mathbf{x}^{(1:L)}_{\text{AIS}}$ using AIS, with $p(\mathbf{x})$ as the target distribution and $q_\theta(\mathbf{x})$ as the proposal distribution. 
We can then obtain an importance weighted estimate of the above gradient operator
\begin{equation}
\label{eqn:un_log_alpha_div_MC}
    \operatorname{E}_{p(\mathbf{x})} \left[\nabla_\theta  \frac{p(\mathbf{x})}{q_{\theta}(\mathbf{x})}\right] \underset{\sim}{\propto}\sum_{l=1}^L \bar{w}^{(l)}_{\text{AIS}} \left[ \nabla_\theta \frac{p(\bar{\mathbf{x}}^{(l)}_{\text{AIS}})}{q_\theta(\bar{\mathbf{x}}^{(l)}_{\text{AIS}})}\right].
\end{equation}
We note that in Equation \ref{eqn:gradient_estimator_alpha_div}, $q_\theta(\mathbf{x})$ is only differentiated through $\theta$'s contribution to the probability density function, and not\footnote{In Equation \ref{eqn:gradient_estimator_alpha_div}, $\nabla_\theta$ is inside the expectation, with $\x \sim p(\mathbf{x})$ independent to $\theta$.} via $\nabla_\theta \x$. 
Therefore, in Equation \ref{eqn:un_log_alpha_div_MC} we take care to block the gradient of $\x^{(1:L)}_{\text{AIS}}$ with respect to $\theta$. 
We denote the blocked gradients with $\bar{\x}_{AIS}^{(l)}$. 
Thus, we can train the proposal by minimising the surrogate “loss function” 
\begin{equation}
\begin{aligned}
\label{eqn:surrogate_non_log_alpha_div_loss}
    \mathnormal{O} (\theta) = \sum_{l=1}^L \bar{w}^{(l)}_{\text{AIS}} \left[ \frac{p(\bar{\mathbf{x}}^{(l)}_{\text{AIS}})}{q_\theta(\bar{\mathbf{x}}^{(l)}_{\text{AIS}})}\right], \\
\end{aligned}
\end{equation}
taking care\footnote{In Equation \ref{eqn:un_log_alpha_div_MC}, $w^{(1:L)}_{\text{AIS}}$ is not differentiated with respect to $\theta$, so we must block the gradient of $w^{(1:L)}_{\text{AIS}}$ with respect to $\theta$, as otherwise automatic differentiation will result in an incorrect estimate of the gradient. 
This is because the flow model parameters $\theta$ participate in the calculation of $w^{(1:L)}_{\text{AIS}}$ and $x^{(1:L)}_{\text{AIS}}$.} to block the gradient of $\bar{w}^{(1:L)}_{\text{AIS}}$ and $\bar{\mathbf{x}}^{(1:L)}_{\text{AIS}}$ with respect to $\theta$. 

To obtain a good loss function for training it is beneficial to instead seek to write the surrogate loss (Equation \ref{eqn:surrogate_non_log_alpha_div_loss}) in terms of log probabilities and log importance weights, because inside the expectation the importance weights and fractions of probabilities will have high variance. 
To do this we can re-write the surrogate loss as

\begin{alignat}{3}
\label{eqn:log_alpha_div_MC}
    \mathnormal{O} (\theta) \ &= \exp \log  \sum_{l=1}^L \bar{w}^{(l)}_{\text{AIS}} \bigg[ &&   \frac{p(\bar{\mathbf{x}}^{(l)}_{\text{AIS}})}{q_\theta(\bar{\mathbf{x}}^{(l)}_{\text{AIS}})} \bigg] \nonumber \\
    &=  \exp \log  \sum_{l=1}^L \exp \bigg( &&  \log \bar{w}^{(l)}_{\text{AIS}} +  \nonumber \\
    &  && \left(\log p(\bar{\mathbf{x}}^{(l)}_{\text{AIS}}) - \log q_\theta(\bar{\mathbf{x}}^{(l)}_{\text{AIS}})\right)\bigg). 
\end{alignat}
Finally, we instead minimise $\log \mathnormal{O} (\theta)$,
\begin{equation}
\label{eqn:Boostrap_alpha_div_final_loss}
\begin{aligned}
    \mathcal{L} (\theta) \ = 
    \log  \sum_{l=1}^L \exp \bigg( & \log \bar{w}^{(l)}_{\text{AIS}} \ + \ \\
   & \left(\log p(\bar{\mathbf{x}}^{(l)}_{\text{AIS}}) - \log q_\theta(\bar{\mathbf{x}}^{(l)}_{\text{AIS}})\right) \bigg). \\
\end{aligned}
\end{equation}
which allows us to work with $\log$ probabilities and $\log$ importance weights, and use the “$\operatorname{logsumexp}$” trick to obtain a numerically stable estimate.
Minimising $\log \mathnormal{O} (\theta)$ instead of $\mathnormal{O} (\theta)$ introduces bias (as a result of Jensen's inequality), which decreases as $L$ increases.
Equation \ref{eqn:Boostrap_alpha_div_final_loss} is the exact surrogate loss implemented in practice for training.

We may also extend the FAB approach to alternative training objectives that are in the form of expectations over $p(\x)$.
For example we may wish to minimise the forward KL divergence as an objective using,
\begin{equation}
\begin{aligned}
    \operatorname{KL} \left(p || q \right) &\propto - \operatorname{E}_{p(\mathbf{x})}\left[\log q_\phi(\x) \right] \\
    & \approx \sum_{l=1}^L - \frac{\bar{w}^{(l)}_{\text{AIS}}}{\sum_{l=1}^L \bar{w}^{(l)}_{\text{AIS}}} \log q_\theta(\bar{\mathbf{x}}^{(l)}_{\text{AIS}}). 
\end{aligned}
\end{equation}

\end{document}